\documentclass[10pt,twocolumn,letterpaper]{article}

\usepackage[pagenumbers]{cvpr} % To force page numbers, e.g. for an arXiv version

\usepackage{graphicx}
\usepackage{amsmath}
\usepackage{amssymb}
\usepackage{booktabs}
\usepackage{algorithm}
\usepackage{algorithmic}
\usepackage{multirow}
\usepackage{multicol}
\usepackage[pagebackref,breaklinks,colorlinks]{hyperref}
\usepackage{textcomp}

\usepackage[capitalize]{cleveref}
\crefname{section}{Sec.}{Secs.}
\Crefname{section}{Section}{Sections}
\Crefname{table}{Table}{Tables}
\crefname{table}{Tab.}{Tabs.}

\def\cvprPaperID{5188} % *** Enter the CVPR Paper ID here
\def\confName{CVPR}
\def\confYear{2023}
\newcommand{\pparagraph}[1]{\vspace{1pt} \noindent \textbf{#1} }

\begin{document}

%%%%%%%%% TITLE - PLEASE UPDATE
\title{HGFormer: Hierarchical Grouping Transformer for Domain Generalized Semantic Segmentation}

% \author{Jian Ding\\
% Institution1\\
% Institution1 address\\
% {\tt\small firstauthor@i1.org}
% % For a paper whose authors are all at the same institution,
% % omit the following lines up until the closing ``}''.
% % Additional authors and addresses can be added with ``\and'',
% % just like the second author.
% % To save space, use either the email address or home page, not both
% \and
% Second Author\\
% Institution2\\
% First line of institution2 address\\
% {\tt\small secondauthor@i2.org}
% }

\author{
	% Jian Ding, Nan Xue, Gui-Song Xia\thanks{Corresponding author: guisong.xia@whu.edu.cn.}, Dengxin Dai \\
	Jian Ding$^{1,2,3}$, Nan Xue$^{1}$, Gui-Song Xia$^{1, 2}$\thanks{Corresponding author}, Bernt Schiele$^{3}$, 
	Dengxin Dai$^{3}$\\
	$^{1}$NERCMS, School of Computer Science, Wuhan University, China \\
	$^{2}$State Key Lab. LIESMARS, Wuhan University, China \\
  $^{3}$Max Planck Institute for Informatics, Saarland Informatics Campus, Germany\\
	{\tt\small \{jian.ding, xuenan, guisong.xia\}@whu.edu.cn, \{schiele, ddai\}@mpi-inf.mpg.de}
}

\maketitle

% Different from the previous works, which usually study DG for segmentation in the aspect of normalization, we study the DG for segmentation from the aspect of segmentation formulation. First, we give a general view of semantic segmentation formulation, which is a classification task on image partitions.
%%%%%%%%% ABSTRACT
\begin{abstract}
Current semantic segmentation models have achieved great success under the independent and identically distributed (i.i.d.) condition. However, in real-world applications, test data might come from a different domain than training data. Therefore, it is important to improve model robustness against domain differences.  
This work studies semantic segmentation under the domain generalization setting, where a model is trained only on the source domain and tested on the unseen target domain.
Existing works show that Vision Transformers are more robust than CNNs and show that this is related to the visual grouping property of self-attention. In this work, we propose a novel hierarchical grouping transformer (HGFormer) to explicitly group pixels to form part-level masks and then whole-level masks. The masks at different scales aim to segment out both parts and a whole of classes. HGFormer combines mask classification results at both scales for class label prediction. 
%Since grouping is naturally aligned with image segmentation, we develop a novel transformer network that can explicitly group pixels into masks and apply mask classification for class label prediction to increase robustness. Existing works such as Mask2former     
%a few recent semantic segmentation methods (\eg Mask2former\cite{mask2former}) 
%it is of great interest to develop novel semantic segmentation networks that have explicit grouping operations to improve robustness. Recently,
%Intuitively, making classifications on grouped masks is more robust than classifications on pixels, since masks have larger receptive fields. 
%However, we find the process to group pixels into whole-level masks is not as robust as grouping pixels into part-level masks. Therefore, we designed a hierarchical grouping transformer (HGFormer) to efficiently generate both part-level and whole-level masks. The final semantic segmentation results are the ensemble of mask classifications at different scales. 
We assemble multiple interesting cross-domain settings by using seven public semantic segmentation datasets. Experiments show that HGFormer yields more robust semantic segmentation results than per-pixel classification methods and flat-grouping transformers, and outperforms previous methods significantly. Code will be available at \url{https://github.com/dingjiansw101/HGFormer}.

\end{abstract}
% 	\caption{Image partitions at different scales. Most deep learning based semantic segmentation methods make classification on low-level grid cells. Some recent works make classification on whole-level masks (which are close to the ground truth masks). The images can also be partitioned into part-level masks. We propose to make classification on both part-level and whole-level for robust semantic segmentation.  
% 	}
%%%%%%%%% BODY TEXT

% 	The classification units can be roughly divided into three categories: low-level grid cells, part-level masks, and whole-level masks. 
% In contrast, the low-level  grid cell partitioning is fixed under different domains, but the size of the classification unit is small, which makes the classification not reliable. 

\section{Introduction}
\label{sec:intro}
\begin{figure}[!t]
	\centering
% 	\vspace{-3mm}
	\includegraphics[width=0.99\linewidth]{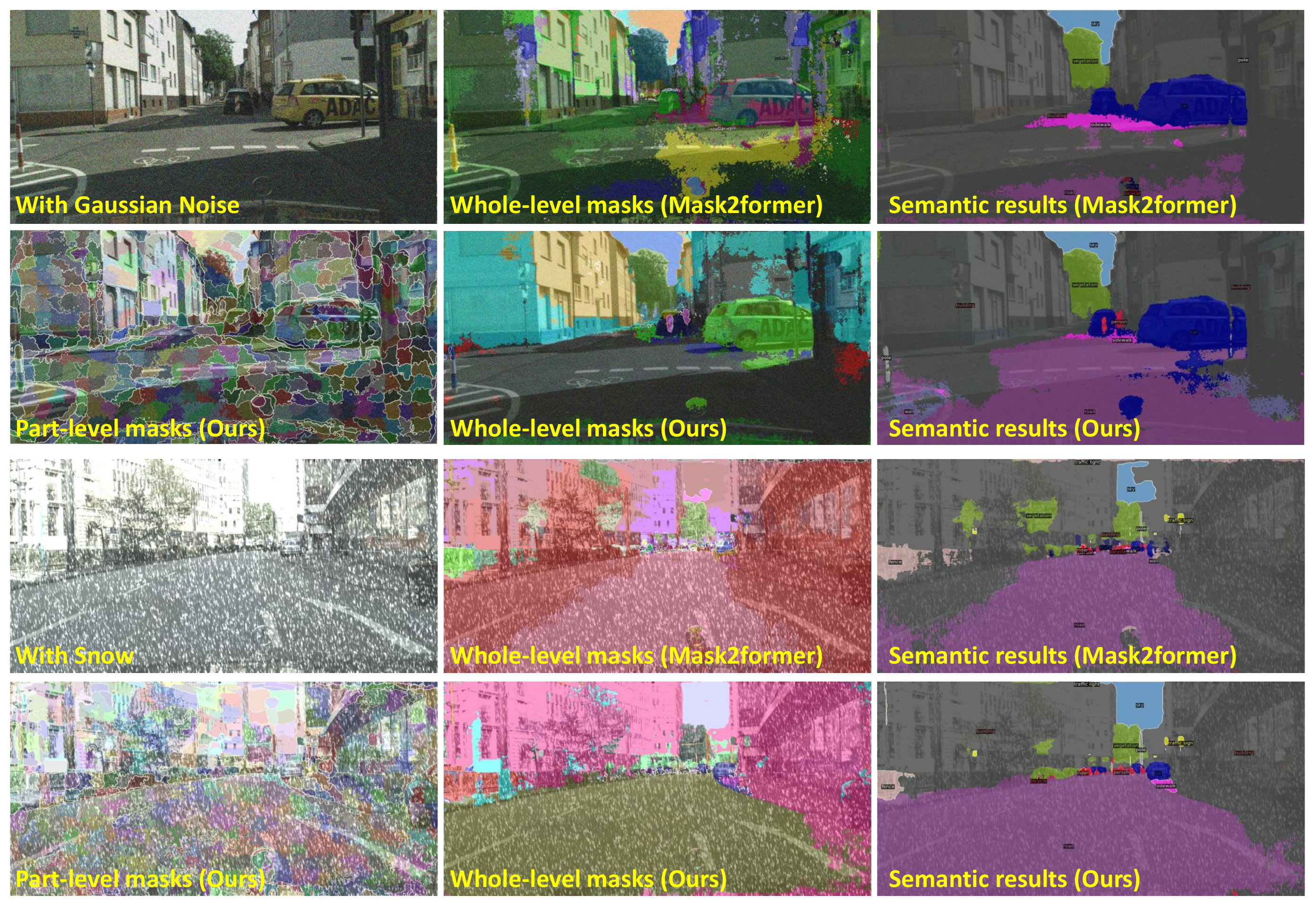}
	\vspace{-3mm}	
	\caption{Semantic segmentation can be considered as partitioning an image into classification units (regions), then classifying the units. The units can range from pixels to large masks. Intuitively, mask classification is more robust than per-pixel classification, as \textit{masks} allow to aggregate features over large image regions of the same class to predict a `global' label. Despite this promise, the process of grouping pixels into \textit{whole-level} masks directly from pixels is very challenging under the distribution shift (\eg, Gaussian Noise). In order to tackle this problem, we present a hierarchical grouping paradigm to group pixels to \emph{part-level masks} first and then to group \emph{part-level masks} to \emph{whole-level masks} to get reliable masks. Then we combine both part-level and whole-level mask classification for robust semantic segmentation, given that the masks at the two levels capture complementary information.       
	}
	\label{fig:partitions}
	\vspace{-3mm}
\end{figure}
% Intuitively, mask classification is more robust than per-pixel classification, as \textit{masks} allow to aggregate features over large image regions of the same class to predict a `global' label. Despite of this promise, generating whole-level masks directly from pixels is a very challenging task.

% In order to tackle this problem, this work proposes using hierarchical grouping in the segmentation transformer architecture to group pixels to \emph{part-level masks} first and then to group \emph{part-level masks} to \emph{whole-level masks}.

% The part-level partitioning and classification is a more robust partitioning than the whole-level partitioning, although it has smaller area of regions. Therefore, they are complementary.  We use both part-level and whole-level mask classification for semantic segmentation.

Research in semantic image segmentation has leaped forward in the past years due to the development of deep neural network. However, most of these models assume that the training and testing data follow the same distribution. In the real-world, we frequently encounter testing data that is out of distribution. The generalization ability of models under distribution shift is crucial for applications related to safety, such as self-driving.  
In domain generalization setting, models are trained only on source domains and tested on target domains, where the distributions of source domains and target domains are different. Unlike the domain adaptation~\cite{DDA,hoyer2022daformer},  target data is not accessible / needed during training, making the task challenging but practically useful. 

% Although there are many works on domain generalized classification. 
% There have been a few works on domain generalization (DG) for semantic segmentation. They tackled DG for semantic segmentation from the perspectives of domain randomization~\cite{GTR}, normalization and whitening~\cite{ISW, IBN, SW, SAN}. Beyond the segmentation task, an important line of works for out-of-distribution robustness is model architecture designs. 

% Although not designed specifically for robustness and generalization, 
Recently, Vision Transformers have been shown to be significantly more robust than traditional CNNs in the out-of-distribution generalization~\cite{segformer,FAN,vitrobust,assyingood,hoyer2022daformer,guo2023improving}. 
% The self-attention in Vision Transformer was claimed to be crucial for improving robustness~\cite{vitrobust}. 
Some works interpret self-attention as a kind of \textit{visual grouping}~\cite{emergingvit,intriguingvit}, and believe that it \textit{is related to robustness}~\cite{FAN}. However, these works mainly focus on classification. Although FAN~\cite{FAN} and Segformer~\cite{segformer} have been evaluated on segmentation, they do not explicitly introduce visual grouping in their networks. Since grouping is naturally aligned with the task of semantic segmentation, we would like to ask the question: \textit{can we improve the robustness of semantic segmentation by introducing an explicit grouping mechanism into semantic segmentation networks?}

Most deep learning based segmentation models directly conduct per-pixel classification without the process of grouping. Some recent segmentation models introduced \textit{flat grouping}~\cite{maskformer,kmasktransformer} into the segmentation decoder, where pixels are grouped into a set of binary masks directly and classification on masks is used to make label prediction. By using a one-to-one matching similar to DETR~\cite{detr}, the loss between predicted masks and ground truth masks is computed. Therefore the network is trained to directly predict \textit{whole-level masks}, as shown in Fig.~\ref{fig:partitions}. Intuitively, if whole-level masks are accurate, mask classification will be more robust than per-pixel classification due to its information aggregation over regions of the same class. But we find that using the \emph{flat grouping} to generate whole-level masks is susceptible to errors, especially under cross-domain settings. This is shown by the example in Fig.~\ref{fig:partitions} - bottom. 

Different from the \textit{flat grouping} works~\cite{mask2former,kmasktransformer}, we propose a \textit{hierarchical grouping} in the segmentation decoder, where the pixels are first grouped into \textit{part-level masks}, and then grouped into \emph{whole-level masks}. Actually, the hierarchical grouping is inspired by the pioneer works of image segmentation~\cite{contour,imageparsing,Chen_2016_CVPR} and is further supported by strong psychological evidence that humans parse scenes into part-whole hierarchies~\cite{hinton2021represent}. We find that grouping pixels to part-level masks and then to whole-level masks is more robust than grouping pixels directly to whole-level masks. Part-level masks and whole-level masks segment images at different scales such as \emph{parts} and \emph{a whole} of classes. Therefore, part-level and whole-level masks are complementary, and combining mask classification results at those different scales improves the overall robustness. %and react differently to the domain shifts. The combination of classification on different levels can further improve the generalization. 

% Different from the previous works, we study the domain generalized semantic segmentation models in the aspect of segmentation formulations. First, we categorize the formulation of semantic segmentation into pixel-level classification~\cite{FCN,Deeplab,Deeplabv2,Deeplabv3,Deeplabv3plus}, part-level mask classification~\cite{regionproxy}, and whole-level mask classification~\cite{maskformer,mask2former,maxdeeplab,kmasktransformer,cmtdeeplab,KNet}. These different types of semantic segmentation can all be considered as partitioning an image into regions then classify regions. The difference is at the scale of regions, which can be viewed in Fig.~\ref{fig:partitions}. 
% Actually, these works already used the mask classification; however, the one-to-one matching loss~\cite{detr} forces the model to predict the region at the largest scale.

% FAN interprete the robustness from the aspect of grouping. information bottleneck, glome, hierchical parsing is more interpretable. spix is a kind of dynamic structure, which can adaptively adjust to the target images

% In this paper, we generate the regions at different scales. We also generate different partitions at the same scale.  
% In this paper, we generate a diverse set of partitions. Then we classify these regions. We hypothesis each partition sub-module reacted differently to the \textit{domain shift}. When we make classifications on different types of partitions and scales, we can make more robust semantic segmentation.

To instantiate a hierarchical grouping idea, we propose a hierarchical grouping transformer (HGFormer) in the decoder of a segmentation model. The diagram is shown in Fig.~\ref{fig:pipeline}. We first send the feature maps to the part-level grouping module. In the part-level grouping module, the initialization of cluster centers is down sampled from feature maps. Then we compute the \textit{pixel-center similarities} and assign pixels to cluster centers according to the similarities. To get the part-level masks, we only compute the similarities between each pixel feature and its nearby center features. We then aggregate information of the part-level masks and generate whole-level masks by using cross-attention, similar to how previous methods aggregate pixels information to generate whole-level masks~\cite{mask2former,kmasktransformer}. Finally, we classify masks at different levels, and average the semantic segmentation results of all the scales.

We evaluate the method under multiple settings, which are assembled by using seven challenging semantic segmentation datasets. In each of the setting, we train the methods on one domain and test them on other domains. Extensive experiments show that our model is significantly better than previous \textit{per-pixel classification based}, and \textit{whole-level mask based} segmentation models for out-of-distribution generalization.

% To achieve this, we design a hierarchical grouping model, which can efficiently generate diverse partitions at different scales. Firstly, we perform an iterative superpixel learning process, which groups the pixels into superpixels at the part-level. Then these superpixels are grouped into final whole-level masks. Classification at different partitions and scales can generate semantic segmentation separately. The final semantic segmentation results combine each individual semantic segmentation result. We conduct the experiments on 7 semantic segmentation datasets, which covers diverse domains. Our proposed method is shown to be more robust than the previous segmentation models.

To summarize, our contributions are: 1) We present a hierarchical grouping paradigm for robust semantic segmentation; 2) based on the hierarchical grouping paradigm, we propose a hierarchical grouping transformer (HGFormer), where the pixels are first grouped into part-level masks, and then grouped into whole-level masks. Final semantic segmentation results are obtained by making classifications on all masks; 3) HGFormer outperforms previous semantic segmentation models on domain generalized semantic segmentation across various experimental settings. We also give detailed analyses of the robustness of grouping-based methods under distribution shift.

\section{Related Work}
\subsection{Semantic Segmentation}
% old methods
Semantic segmentation is a classic and fundamental problem in computer vision. It aims to segment the objects and scenes in images and give their classifications.
In the deep learning era, semantic segmentation is usually formulated as a pixel-level classification problem~\cite{FCN,segformer,Deeplab,Deeplabv2,Deeplabv3,Deeplabv3plus} since FCN~\cite{FCN}. Recently, it is becoming popular to use whole-level mask classification to formulate the semantic segmentation problem~\cite{kmasktransformer,maxdeeplab,cmtdeeplab,mask2former,maskformer}. In contrast to pixel-level and mask-level classification, to our best knowledge, there are very few works on learning part-level masks~\cite{SSN,superpixelFCN}, and using part-level mask classification~\cite{regionproxy,superpixelbi} for semantic segmentation in deep learning era. Among them, SSN~\cite{SSN} and super pixel FCN~\cite{superpixelFCN} mainly focus on part-level mask learning instead of part-level classification for the semantic segmentation results. BI~\cite{superpixelbi} is not an end-to-end model, which needs extra part-level masks as input. RegProxy~\cite{regionproxy} is a recent work that closes to our work, which uses convolutions to learn part-level masks, and is only evaluated on the i.i.d. condition. In contrast, we use similarity-based grouping to learn part-level masks, and are the first to validate the effectiveness of using part-level mask classification for domain generalized semantic segmentation. Besides, Regproxy~\cite{regionproxy} is customized with \textit{plain} ViT~\cite{vit}, while our work is applicable to pyramid transformers~\cite{swin,pvt} and CNNs. Our work is also different for the hierarchical segmentation design.
% \subsection{Hierarchical representation}
% say something about the glome

% TODO: say something about strength weak subnets, and MoE

% \subsection{Domain Adaptation}
% Domain adaptation (DA)~\cite{DDA} aims to adapt models trained on source domains to target domains, where source and target domains have different distributions. Most of the previous works focused on the unsupervised domain adaptation (UDA)~\cite{UDA}, where only the labels of source domain is available. Methods for UDA can be divided into adversarial training~\cite{Dlow,fcnwild} and self-training approaches~\cite{ModelAdapt,zou2019confidence}. In the self-training methods, the models are first trained on source domain then used to compute the pseudo labels on the target domain. Based on the pseudo labels, the models trained re-trained. The process is repeated. 
% TODO: extend this part.

\subsection{Domain Generalization}
% Unlike domain adaptation(DA), domain generalization (DG) assumes that the target data (even unlabelled) is not accessible during training, which makes the task much more challenging. 
Domain generalization (DG) assumes that the target data (even unlabelled) is not accessible during training. 
Methods for DG in classification includes domain alignment~\cite{muandet2013domain,li2018domain}, meta-learning~\cite{li2018learning,balaji2018metareg}, data augmentation~\cite{volpi2019addressing,xu2020robust}, ensemble learning~\cite{liu2020ms,cha2021domain,crossensemble}, self-supervised learning~\cite{bucci2021self,carlucci2019domain}, and regularization strategies~\cite{wang2019learning,huang2020self}. 
% DG can be categorized into multi-source DG~\cite{} and single-source DG~\cite{}, according to the number of domains used during training. Most of the previous works studied the multi-source DG.
The ensemble of mask classification at different levels is related to the ensemble methods for domain generalization. The drawback of the previous ensemble-based methods~\cite{yeo2021robustness,lakshminarayanan2017simple} is that they will largely increase the runtime. Some ensemble methods~\cite{wortsman2022model, rame2022diverse} focus on the averaging of model weights, which do not increase the runtime, but increase the training time. Our method does not introduce extra FLOPS due to the efficient hierarchical grouping design, and does not introduce extra training time.

While the DG in classification is widely studied in the previous works, there are only several works that study the DG in semantic segmentation. The previous methods for DG in semantic segmentation includes: (1) Domain Randomization~\cite{DRPC, GTR} and (2) Normalization and Whitening~\cite{ISW, IBN, SW, SAN}.
Although not designed specifically for domain generalization tasks, the Vision Transformers~\cite{dosovitskiy2020image} have shown their robustness~\cite{FAN} in the out-of-distribution setting. The robustness of Vision Transformer was explained to be related to the grouping property of self-attention~\cite{FAN}. However, there are no works that study the effect of explicit grouping in the segmentation decoder for semantic segmentation in DG. Motivated by these results, we study the different levels of grouping and mask classification for semantic segmentation in DG.

\begin{figure*}[!t]
	\centering
	\vspace{-3mm}
	\includegraphics[width=0.86\linewidth]{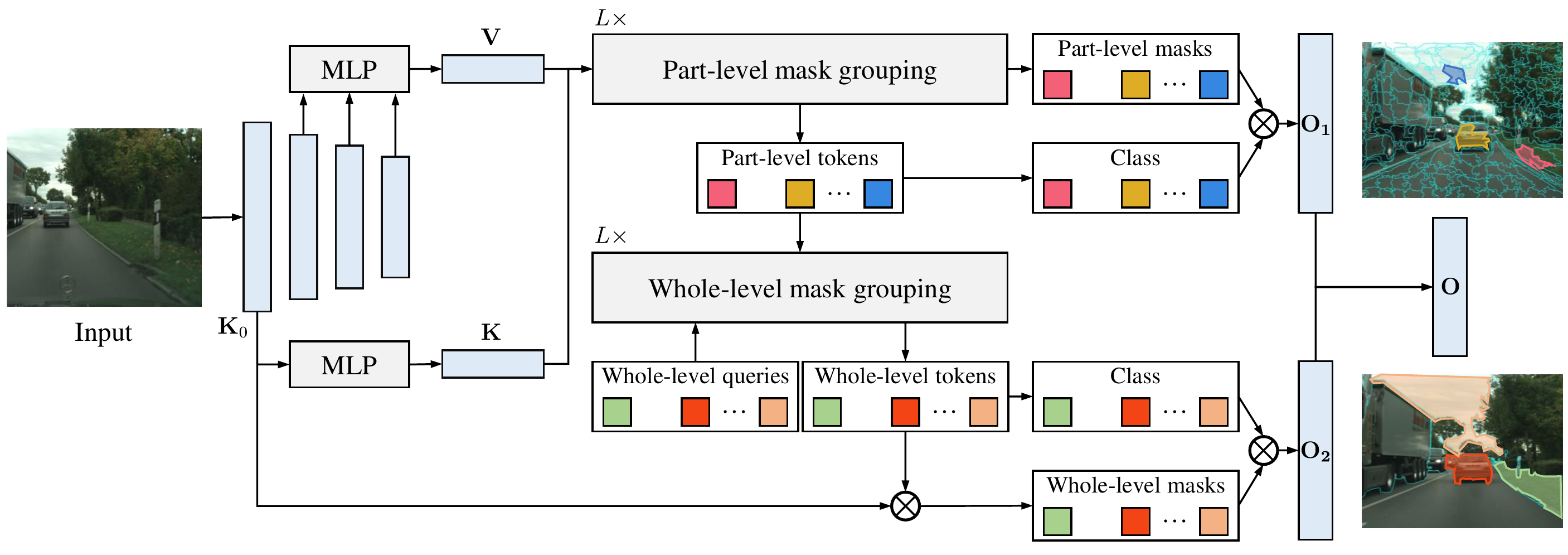}
	\vspace{-3mm}	
	\caption{The pipeline of our proposed method. We first pass an image to a backbone network and get feature maps at different resolutions. The largest feature map $\mathbf{K}_0$ is projected to $\mathbf{K}$ for part-level grouping. The other three feature maps are fused to form a new feature map $V$ for part-level mask feature extraction used for later classification. The details of part-level grouping can be seen in Algorithm~\ref{alg:training}. The grouping process is repeated $L$ iterations. At the end of each iteration, there are $N_p$ part-level masks, and their tokens. Combining part-level classifications and part-level masks, we can get the semantic segmentation results $\mathbf{O}_1$. The part-level tokens from the last iteration of part-level grouping are aggregated to whole-level masks by whole-level grouping (which are actually cross-attention layers). Similarly, there are also $L$ iterations in the whole-level grouping. At the end of each iteration, there are $N_o$ whole-level tokens. Whole-level masks are computed by a matrix multiplication between $\mathbf{K}_0$ and projected whole-level mask tokens. Similarly, we can get semantic segmentation results $\mathbf{O}_2$ by combining whole-level masks and their classifications. The final results $\mathbf{O}$ are the sum of $\mathbf{O}_1$ and $\mathbf{O}_2 $.} 
	\label{fig:pipeline}
	\vspace{-3mm}
\end{figure*}

\section{Methods}
%\bernt{I am lacking an overview of the method at the beginnging of this section - also a concept figure would be great to discuss right here. The only thing the reader knows so far that we want to do something hiearchical - but not why and how this is done. It would be good to spend at least a paragraph to give an overview of the overall method including the reason for the hiearchical approach and also what else the approach contains right here -- sec 3.1 already defines low- mid- and whole-level masks and it is not even clear why that could be interesting to consider} 

% \jian{Semantic segmentation results can be split into two steps: 1) partitioning an image into regions; and 2) classifying the regions.  As can be seen in Fig.~\ref{fig:partitions},}
% In this section, we first review previous pixel based, whole-level mask based, and part-level mask based segmentation models. They can all be formulated by one unified formulation. Then we discuss our proposed hierarchical segmentation model as an instantiation thereof.

%\bernt{the intro paragraph does not really make sense to me - if I understand correctly, the classic levels are either pixel-level or whole-level -- the novelty and difference in this work is that we also have the part-level -- right? if that is the case we should probably right exactly this and then the flow of the following paragraphs would be more logical to me. Does that make sense?} 

%\subsection{A general view of semantic segmentation}
 Given an image $I \in \mathbb{R}^{H\times W \times 3}$, an image partition is defined as $S=\{R_1, ..., R_N\}$, such that  $\cup^N_{i=1} R_i = \Omega$ and $R_i \cap R_j = \O$, if $i \neq j$. After mapping each region $R_i$ to a class by $L_i$, we get $C=\{L_1(R_1), ..., L_N(R_N)\}$. Semantic segmentation can then be defined as:
\begin{equation}
\label{seg} 
Y = \{S, C\}.
\end{equation}
% \begin{figure*}[!t]
% 	\centering
% 	\includegraphics[width=0.97\linewidth]{images/partitionsv2.pdf}
% 	\vspace{-4mm}	
% 	\caption{Explanation of the association between pixels and its nearby grid cells. For each pixel in green box, we compute the similarities between the pixel and its 9 nearby regular grid cells.  
% 	}
% 	\label{fig:spixgrid}
% 	\vspace{-2mm}
% \end{figure*}
According to the scales, we can roughly divide the image partitions into three levels: pixels, part-level masks, and whole-level masks. The \textit{pixel} partition, where each element in $S$ is a pixel, is widely used by all per-pixel classification methods \cite{FCN,Deeplabv2}. The \textit{whole-level masks} partition, where each element in $S$ represents a whole mask of a class, is used by a few recent approaches \cite{maskformer,mask2former}. The \textit{part-level mask} partition, where each element aims to cover a class part, is proposed by this work and used along with \emph{whole-level masks} to enhance robustness.

Intuitively, mask classification is more robust than per-pixel classification, as \textit{masks} allow to aggregate features over large image regions of the same class to predict a `global' label. Despite of this promise, generating whole-level masks directly from pixels is a very challenging task. While SOTA methods~\cite{maskformer,mask2former} can generate reasonable \emph{whole-level masks} directly from pixels, the \emph{flat grouping} methods used are not robust to domain changes -- when tested on a different domain, the generated masks are of poor quality, leading to low semantic segmentation performance (see Fig.~\ref{fig:partitions}). In order to tackle this problem, this work proposes using hierarchical grouping in the segmentation transformer architecture to group pixels to \emph{part-level masks} first and then to group \emph{part-level masks} to \emph{whole-level masks}. The advantages are 1) the grouping task becomes less challenging; 2) mask classification can be performed at both scales and their results can be combined for more robust label prediction, given that the \emph{masks} at the two scales capture complementary information; 3) global self-attention to aggregate long-range context information can be performed directly at both part-level and whole-level now, which is not possible at pixel-level due to the huge computation cost.

%The \textit{part-level mask classification can be considered as a trade off} between pixel-level classification and whole-level mask classification. We find that our proposed part-level grouping is more robust than the previous whole-level grouping~\cite{mask2former} \bernt{I guess we need refernces of related work here that does whole-level grouping - and those should be ideally include the works we compare to later}. And the generated part-level masks (by grouping) have larger receptive fields than the low-level grid cells. We can also perform self-attention to aggregate the global information on part-level masks, which is not possible on low-level grid cells due to the huge computation brought by the increased resolution. This is different from previous methods which only give predictions on a single level. Our method classifies both at part-level and whole-level masks. Then we combine the classification results to get the final semantic segmentation result.

% TODO: refer freesolo, super pixel ssn, super pixel fcn, slic
% TODO: check the details of algorithm
\begin{algorithm}[t]
\caption{Part-level grouping
}
% =\frac{ \exp(\textbf{D}_{i,j})}{\sum_{i=1}^{N_p}\exp(\textbf{D}_{i,j})}
% TODO: polish the algorithm
\label{alg:training}
    \begin{algorithmic}[1]
    	\REQUIRE
    	\textit{Pixel} feature map $\mathbf{K}\in \mathbb{R}^{(H\times W)\times d}$, classification feature map $\mathbf{V}\in \mathbb{R}^{(H\times W)\times d}$ \\
    	\STATE Initialize the cluster \textit{center} features $\mathbf{Q^1}\in \mathbb{R}^{N_p\times d}$ by down sampling $\mathbf{K}$
        \FOR{$t=1, \cdots, L$}
        \STATE Compute assignment matrix $\mathbf{A^t}$ by $\mathbf{Q^t}$ and $\mathbf{K}$  
        \STATE Update the cluster center features $\mathbf{Q^{t+1}} = \mathbf{A^t}\times \mathbf{K}$
        
        \STATE Update the part-level tokens $\mathbf{Z^{t}} = \mathbf{A^t}\times \mathbf{V}$
                
        \ENDFOR
    \end{algorithmic}
\end{algorithm}

\subsection{HGFormer}
To efficiently implement a model which can predict semantic segmentation at different scales, we adopt a hierarchical grouping process, which consists of two stages. The first is to group pixels into part-level masks by similarity-based local clustering. The second is to group part-level masks into whole-level masks by cross-attention. Then we make classification on partitions at different scales. The framework can be seen in Fig.~\ref{fig:pipeline}. We will introduce the details in the following. 

\pparagraph{Part-level grouping.} The goal of part-level grouping is to compute a partition $S_m=\{R_1,...,R_{N_p}\}$, with $N_p$ the number of part-level masks. $S_m$ can be represented by a \textit{hard assignment} matrix $\tilde{\mathbf{A}}\in \{0, 1\}^{N_p\times (HW)}$ such that $\tilde{\mathbf{A}}_{ij}=1$ if the $j$-th pixel is assigned to mask $i$ and 0 otherwise. Since the hard assignment is not differentiable, we compute a \textit{soft assignment} matrix $\mathbf{A}\in [0, 1]^{N_p\times (HW)}$ such that $\sum_{i=1}^{N_p}\mathbf{A}_{ij}=1$. $\mathbf{A}_{ij}$ represents the probability of assigning the $j$-th pixel to mask $i$.

\begin{figure}[!t]
	\centering
	\includegraphics[width=0.45\linewidth]{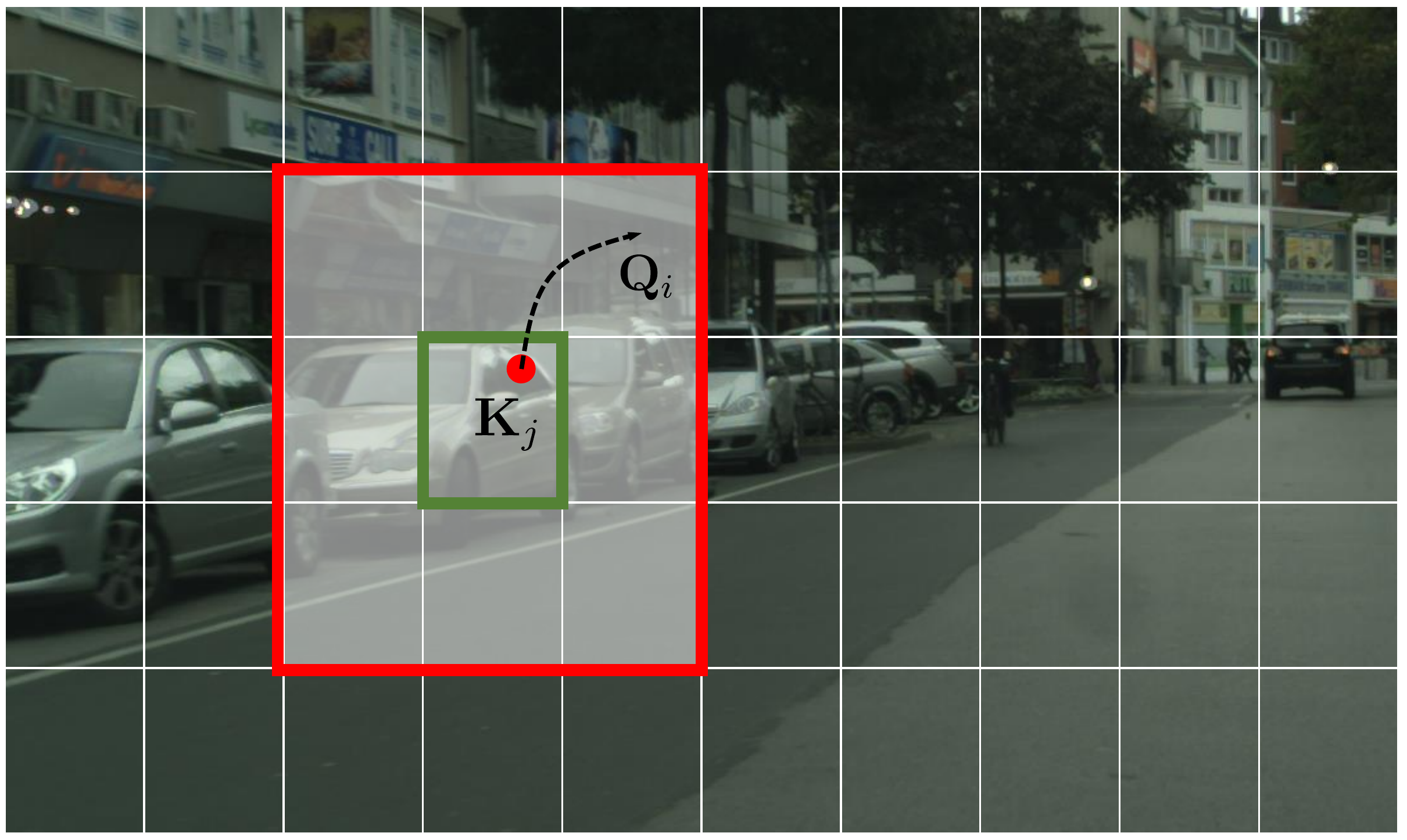}
	\vspace{-2mm}	
	\caption{Explanation of the similarities between pixel features and its nearby center features. The grouping process is to assign each pixel to one of $N_p$ center features. However, due to the computation cost of the \textit{global} comparisons, we only compute the similarities between pixels and their nearby center features to perform \textit{local} comparisons. For example, we only assign each pixel in the green box to one of its 9 nearby center features.
	}
	\label{fig:spixgrid}
\end{figure}

To compute $\mathbf{A}$, we perform an iterative grouping algorithm (see Algorithm~\ref{alg:training}). It takes a feature map $\mathbf{K}\in \mathbb{R}^{(H\times W) \times d}$ as input to compute \textit{assignment matrix}.
% Passing an image to a backbone network (CNN~\cite{resnet} or Transformers~\cite{vit, swin, pvt}), we get two feature maps, $\mathbf{K}\in \mathbb{R}^{(H\times W) \times d}$ and $\mathbf{V} \in \mathbb{R}^{H\times W \times d}$, used for \textit{assignment matrix computation}, respectively. 
We partition an image feature map $\mathbf{K}$ into regular grid cells with size $(r\times r)$ as the initialization of part-level masks, which is a common strategy in super pixel learning~\cite{SSN,SLIC,superpixelFCN}. Then we average the features inside regular grid cells to get the features of part-level masks (or called cluster center) features $\mathbf{Q}\in \mathbb{R}^{N_p \times d}$, where $N_p=H/r\times W/r$. Then we compute the cosine similarities between pixel-center pairs and get $\mathbf{D}\in \mathbb{R}^{N_p\times (HW)}$. For efficiency, we do not compute the similarities between all pixel-center pairs. Instead, we only compute the similarities between pixels and their 9 nearby centers (see Fig.~\ref{fig:spixgrid}). As a result, we get $\mathbf{D}'\in \mathbb{R}^{9\times (HW)}$. But for the convenience of describing, we still use the $\mathbf{D}$ in the following. Due to the local constraint, each cluster center can only aggregate the nearby pixels, so we can get the \textit{part-level masks}. 

The similarities between the $i$-th center feature and $j$-th pixel feature are written as:
\begin{equation}
\label{eq:affinity} 
\mathbf{D}_{i,j} = \left\{\begin{matrix} f(\mathbf{Q}_i, \mathbf{K}_{j}) & \text{if } ~i \in N_j
\\  -\infty& \text{if } ~i \notin N_j,
\end{matrix}\right.
\end{equation}
where $\mathbf{Q}_i \in \mathbb{R}^{d}$ is the $i$-th cluster center feature, and $\mathbf{K}_j \in \mathbb{R}^{d}$ is the $j$-th pixel feature. $f(\bf \mathbf{x}, \mathbf{y})=\frac{1}{\tau} \frac{\bf \mathbf{x}\cdot \bf \mathbf{y}}{|\mathbf{x}| \cdot |\mathbf{y}|}$ computes the cosine similarity between $\mathbf{x}$ and $\mathbf{y}$, where $\tau$ is the temperature to adjust the scale of similarities. $N_j$ is the set of nearby regular grid cells of $j$-th pixel, which can be viewed in Fig.~\ref{fig:spixgrid}. Then we can compute the soft assignment matrix as:
\begin{equation}
\mathbf{A}_{i,j} = \text{softmax}(\textbf{D})(i,j) = \frac{ \exp(\textbf{D}_{i,j})}{\sum_{i=1}^{N_p}\exp(\textbf{D}_{i,j})}, 
\end{equation}
Then we can update the cluster center features by 
\begin{equation}
\label{updatespixfeature} 
\mathbf{Q}_{new} = \mathbf{A}\times \mathbf{K}.
\end{equation}
After we get the new center features, we can compute the new assignment matrix by using updated center features $\mathbf{Q}_{new}$ and feature map $\mathbf{K}$. The process is repeated $L$ times, as shown in Algorithm~\ref{alg:training}. To get the part-level mask tokens for classification, we use the assignment matrix to extract part-level tokens from another feature map $\mathbf{V}$ by
\begin{equation}
\label{updatespixfeature} 
\mathbf{Z} = \mathbf{A}\times \mathbf{V}.
\end{equation}
To strengthen the part-level mask features, we pass $\mathbf{Z}$ to a self-attention layer and a feed forward network (FFN) layer and get $\mathbf{Z}'$. 
Then we use a linear classifier layer and a softmax activation layer to map $\mathbf{Z}'\in \mathbb{R}^{N_p\times d}$ to part-level class predictions $\mathbf{P}_{m}\in \mathbb{R}^{N_p\times K}$, where $K$ is the number of classes. 

Note that we use different feature maps for part-level grouping and classification to decouple these two kinds of tasks, since the shallow layers are usually used for localization while the deep layers benefit the classification~\cite{fpn}.

% \vspace{.5em}\noindent
% \emph{Relation to cross-attention.} The format of the local clustering is similar to the cross-attention~\cite{}. However, there are several differences.
% \textcolor{blue}{TODO: finish the relation to cross-attention}
% TODO: plot a figure to explain it.

%  =  Softmax(\mathbf{Q_t}\times \mathbf{K}^{T})
\pparagraph{Whole-level grouping.}
The aim of this stage is to group part-level masks into whole-level masks. Our framework is agnostic to the detailed grouping method. But for a fair comparison, we use the transformer decoders for grouping, following~\cite{mask2former}. Each transformer decoder layer consists of one multi-head cross-attention, one self-attention, and an FFN layer. Firstly, we perform cross-attention between $N_o$ learnable positional embeddings $\mathbf{E}\in \mathbb{R}^{N_o\times d}$ and part-level tokens $\mathbf{Z}'\in \mathbb{R}^{N_p\times d}$ to get the output features:
\begin{equation}
\label{updatespixfeature} 
\mathbf{E}_{out} = \text{Softmax}((\mathbf{E}\mathbf{W}_q) \times (\mathbf{Z}'\mathbf{W}_k)^{T})\times (\mathbf{Z}' \mathbf{W}_v ),
\end{equation}
where $W_q\in \mathbb{R}^{d\times d}$, $W_k\in \mathbb{R}^{d\times d}$, $W_v\in \mathbb{R}^{d\times d}$ are projection heads for queries, keys, and values, respectively.
For simplicity, the multi-head mechanism is ignored in the formulation. The cross-attention layer is followed by self-attention, and $\mathbf{E}_{out}$ and FFN layers. Similar to the process in part-level mask learning, the transformer decoder operation is also repeated $L$ times. At the end of each transformer decoder layer, there are two MLP layers. The first MLP layer maps the out features $\mathbf{E}_{out}\in \mathbb{R}^{N\times d}$ to mask embedding $\varepsilon\in \mathbb{R}^{N\times d}$. Then the whole-level masks $\mathbf{M}\in [0, 1]^{N\times (H_0\times W_0)}$ can be computed as
% TODO: check the correctness of sigma formulation 
% TODO: differentiate the K with previous ones
\begin{equation}
\label{sigma}
\mathbf{M} = \sigma(\varepsilon\times \mathbf{K}_0^{T}),
\end{equation}
where $\mathbf{K}_0^{T} \in \mathbb{R}(H\times W)\times d$ is a feature map before $\mathbf{K}$ (see Fig.~\ref{fig:pipeline}). The second MLP layer maps the out features to class logits. Then we apply a softmax activation on the class logits to get the probability predictions $\mathbf{P}_{h}\in \mathbb{R}^{N\times (K+1)}$, where $K$ is the number of classes. There is an extra dimension representing the ``no object'' category ($\varnothing$).

 The difference between our work and the previous counterparts~\cite{mask2former,kmasktransformer} is: the \textit{keys} and \textit{values} in previous works are pixel features, while our \textit{keys} and \textit{values} are features of part-level masks. Our hierarchical grouping design reduces the computation complexity, since the number of part-level masks is much smaller than the pixels.

\pparagraph{Multi-scale semantic segmentation.}
We conduct classification at two levels: part-level mask classification and whole-level mask classification. The semantic segmentation results from the part-level mask classification can be computed as
\begin{equation}
\mathbf{O}_1 = \mathbf{P}_{m}^{T} \times \mathbf{A}
\end{equation}
The semantic segmentation results form the whole-level mask classification can be computed as 
\begin{equation}
\mathbf{O}_2 = \mathbf{P}_{h}^{T} \times \mathbf{M}
\end{equation}
The final semantic segmentation result is an ensemble of two results by a simple addition.
% \begin{equation}
% \mathbf{O} = \mathbf{O_1} + \mathbf{O_2}
% \end{equation}
% Note in the above formulation, we ignored the ``non object" category of $\mathbf{O_2}$, and the resize operation of $\mathbf{O}_1$ and $\mathbf{O}_2$. 

% \subsection{Losses design}
% \pparagraph{losses for part-level mask learning} 
\pparagraph{Loss design.}
The challenge in part-level mask learning is that we do not have the ground truth for the part-level partition. The partition at the part-level stages is not unique. Therefore, we design two kinds of losses. First, we directly add a \textit{cross-entropy} loss $\mathcal{L}_{\text{part,cls}}$ on $\mathbf{O}_1$. Given that the \textit{cross-entropy} loss is also affected by classification accuracy, which does not have a strong constraint on the mask quality, we also add a \textit{pixel-cluster contrastive loss}. The core idea of the contrastive loss is to learn more discriminative feature maps, which can the be used for similarity-based part-level grouping.
Given ground truth masks $\mathbf{M_G}\in \mathbb{R}^{g\times(H\times W)}$, where $g$ is the number of masks in an image, we first average the features within each ground truth mask and get $\mathbf{T}\in \mathbb{R}^{g\times d}$. Then the contrastive loss~\cite{moco,maxdeeplab,van2021unsupervised} for each pixel in $\mathbf{K}$ is computed as:
\begin{equation}
\mathcal{L}_{\text{contrast}}^i = -\log \frac{\sum_{j=1}^{K}\mathbf{M}_{G,j,i} \exp(f(\mathbf{K}_i, \mathbf{T}_j))}{\sum_{j=1}^{K}\exp(f(\mathbf{K}_i, \mathbf{T}_j))},  
\end{equation}
where $f(\mathbf{x}, \mathbf{y})$ is a function to measure the similarity between two feature vectors. 
% The pixel-level contrastive loss has been explored in the previous works~\cite{maxdeeplab,wang2021exploring,detcon,van2021unsupervised}. But as far as we know, we are the first to use it to learn part-level masks. Actually, the contrastive loss is aligned well with Eq.~\ref{eq:affinity}. 
The losses for the whole-level learning mainly follow previous works~\cite{mask2former}. There is a one-to-one matching between predictions and ground truths. For the matched prediction, a dice loss $\mathcal{L}_{\text{dice}}$, a mask loss $\mathcal{L}_{\text{mask}}$, and a mask classification loss $\mathcal{L}_{\text{mask,cls}}$ are computed.
% \begin{equation}
% \mathcal{L} = \lambda_1 \mathcal{L}_{\text{part,cls}} + \lambda_2\mathcal{L}_{\text{contrast}} + \lambda_3\mathcal{L}_{\text{dice}} + \lambda_4\mathcal{L}_{\text{mask}} + \lambda_5\mathcal{L}_{\text{mask,cls}}   
% \end{equation}
% First, we conduct a one-to-one matching between predicted masks $\mathbf{M}$ and $\mathbf{M}_{G}$. 
% It is worthwhile to note that due to the one-to-one matching loss. There will results in many empty masks in $\mathbf{M}$. And the matched masks are forced to be the same as the ground truth masks.

% TODO: describle the detr loss, dice loss, ce loss, matching rule

% \begin{equation}
% \label{updatespixfeature} 
% \mathbf{P} = \text{Softmax}(\mathbf{A})\times \mathbf{V}.
% \end{equation}

\subsection{Implementation Details}
% \pparagraph{hierarchical grouping architecture}

% \pparagraph{Losses weights}
We set the weights for $\mathcal{L}_{\text{part,cls}}$, $\mathcal{L}_{\text{contrast}}$, $\mathcal{L}_{\text{dice}}$, $\mathcal{L}_{\text{mask}}$, and $\mathcal{L}_{\text{mask,cls}}$ to 2, 6, 5, 5, and 2, respectively. The temperature $\tau$ in contrastive loss is 0.1, and the down sample rate $r$ is 4. We set the number of refining stages $L$ in part-level and whole-level grouping to 6. We use the deformable attention Transformer (MSDeformAttn)~\cite{deformabledetr} layers to strengthen the features before we project the feature maps to $\mathbf{K}$ and $\mathbf{V}$. The output strides for $\mathbf{K}$ and $\mathbf{V}$ are 1/8. 
We conduct a multi-scale augmentation and then crop a $512\times 1,024$ patch for training. During testing, we send the images with the original size to models. By default, models are trained with 20k iterations with a batch size of 16. We use the ADAMW as our optimizer
with an initial learning rate of 0.0001 and 0.05 weight decay.
% See more details in our appendix.
\section{Experiments}

\subsection{Datasets}
% We adopt 7 datasets (including Cityscapes~\cite{cityscapes}, BDD~\cite{BDD100k}, Mapillary~\cite{mapillary}, GTAV~\cite{GTAV}, SYNTHIA~\cite{synthia}, Cityscapes-C~\cite{city_c} and ACDC~\cite{ACDC}) to set up our experiments. 

\pparagraph{Real-world datasets.}
Cityscapes contains 5,000 urban scene images collected from 50 cities primarily in Germany. The image size of Cityscapes images is $2,048\times1,024$. BDD is another real-world dataset, which contains 7,000 images for training, and 1,000 images for testing. The images of BDD is mainly collected from US. The image size of BDD is $1,280\times720$. Mapillary~\cite{mapillary} is a large-scale dataset, which contains 18,000 images for training, 2,000 images for validation, and 5,000 images for testing. The images of Mapillary are captured from all over the world, at various conditions regarding weather and season, which makes the dataset very diverse. ACDC \cite{ACDC} collects the images with a resolution of $1,920\times1,080$ under adverse conditions, including night, fog, rain, and snow. ACDC contains 1,600 images for training, 406 images for validation, and 2,000 images for testing. 

\pparagraph{Synthetic datasets.}
GTAV is a synthetic dataset collected from GTAV game, which contains 12,403, 6,382, and 6,181 images with a resolution of $1,914\times1,052$ for training, validation, and testing, respectively. SYNTHIA~\cite{synthia} is another synthetic dataset, which consists of 9,400 photo-realistic images with a size of $1,280\times 760$.

\pparagraph{Common corruption dataset.}
We follow the previous works~\cite{city_c,segformer} to expand the Cityscapes validation set with 16 type of generated corruptions. The corruptions can be divided into 4 categories: noise, blur, weather, and digital. There are 5 severity levels for each kind of corruption.

% TODO
% Cityscapes-C is gnerated from Cityscapes by under the corruptions.

% TODO: extend this part with the number of images.

\subsection{Main Results}
% The models are trained on several dataset and evaluated on other datasets.
% The previous works conduct different experimental settings in the generalization to unseen domains. 
We evaluate models on 4 kinds of generalization settings.

\pparagraph{Normal-to-adverse generalization.}
In this experimental setting, all the models are trained on Cityscapes~\cite{cityscapes} (images at normal conditions) and tested on ACDC~\cite{ACDC} (images at 4 kinds of adverse conditions). Although not specifically designed for domain generalization, transformer-based methods have been shown to be more robust than traditional CNN methods. Therefore, we compare our proposed HGFormer with two representative transformer-based segmentation methods in Tab.~\ref{tab:norm-to-acdc}. Among them, all CNN-based methods and Segformer~\cite{segformer} are based on \textit{per-pixel} classification. Mask2former~\cite{mask2former} is based on \textit{whole-level} classification. We can see that our method outperforms the previous CNN-based methods by a large margin, and also significantly outperforms the competitive transformer-based segmentation models.
\begin{table}[]
\centering
\caption{\textbf{Cityscapes-to-ACDC generalization.} The models are trained on Cityscapes~\cite{cityscapes} only, and tested on ACDC~\cite{ACDC}. The results of Mask2former~\cite{mask2former}, Segformer~\cite{segformer}, and HGFormer are implemented by us. Others are from ACDC paper~\cite{ACDC}.
The results of models trained by us are an average of 3 times. The results of Segformer~\cite{segformer} are obtained by their officially released model. }
\vspace{-3mm}
% \small
\resizebox{\linewidth}{!}{
\begin{tabular}{c|c|cccc|c}
% \hline
\toprule
Method      & backbone & Fog  & Night & Rain  & Snow & All   \\ \hline
RefineNet~\cite{refinenet}   & R101     & 46.4 & 29    & 52.6  & 43.3 & 43.7  \\
DeepLabv2~\cite{Deeplabv2}   & R101     & 33.5 & 30.1  & 44.5  & 40.2 & 38    \\
DeepLabv3+~\cite{Deeplabv3plus}  & R101     & 45.7 & 25    & 50    & 42   & 41.6  \\
DANet~\cite{DANet}       & DA101    & 34.7 & 19.1  & 41.5  & 33.3 & 33.1  \\
HRNet~\cite{hrnet}       & HR-w48   & 38.4 & 20.6  & 44.8  & 35.1 & 35.3  \\ 
Mask2former~\cite{mask2former}  & R50      & 54.1 & 36.5  & 53.1 & 50.6 & 49.8 \\
HGFormer (\textbf{ours})       & R50      & 56.5 & 35.8  & 57.7 & 56.2 & \textbf{53.0} \\ \hline
Mask2former~\cite{mask2former} & Swin-T   & 56.4 & 39.1  & 58.9  & 58.2 & 54.6  \\
Segformer~\cite{segformer}   & B2       & 59.2 & 38.9  & 62.5  & 58.2 & 56.2  \\
HGFormer (\textbf{ours})    & Swin-T   & 58.5 & 43.3  & 62.0  & 58.3 & \textbf{56.7} \\ \hline
Segformer~\cite{segformer}   & B5       & 63.2 & 47.8  & 66.4  & 63.7 & 62.0  \\
Mask2former~\cite{mask2former} & Swin-L   & 69.1 & 53.1  & 68.3  & 65.2 & 65.0 \\
HGFormer (\textbf{ours})    & Swin-L   & 69.9 & 52.7  & 72.0  & 68.6 & \textbf{67.2} \\ 
% \hline
\bottomrule
\end{tabular}\label{tab:norm-to-acdc}
}
\end{table}

\pparagraph{Cityscapes-to-other datasets generalization.} In this experimental setting, models are trained on Cityscapes~\cite{cityscapes} and tested on BDD~\cite{BDD100k}, Mapillary~\cite{mapillary}, GTAV~\cite{GTAV}, and Synthia~\cite{synthia}. The results are shown in Tab.~\ref{tab:city2other}. In the first block of Tab.~\ref{tab:city2other}, we compare all the methods with a ResNet-50~\cite{resnet} backbone. We can see that the grouping-based method Mask2Former~\cite{mask2former} is already comparable to the previous domain generalization methods, which indicates the effectiveness of grouping-based model for generalization. Our HGFormer outperforms Mask2Former~\cite{mask2former} by 1.5 points,  showing that our \textit{hierarchical} grouping-based  model is better than the \textit{flat} grouping-based model for domain generalized semantic segmentation.

\begin{table}[]
\centering
\caption{\textbf{Cityscapes to other datasets generalization.} Models are trained on Cityscapes and tested on BDD (B), Mapillary (M), GTAV (G), and Synthia (S). Results of IBN, SW, DRPC, GTR, ISW, and SAN-SAW are from paper~\cite{SAN}. Others are implemented by us. Our results are an average of 3 times.}
\vspace{-3mm}
% \small
\resizebox{\linewidth}{!}{
\begin{tabular}{c|c|cccc|c}
% \hline
\toprule
    Method        & backbone  & B    & M    & G    & S    & Average \\ \hline
IBN~\cite{IBN}         & R50       & 48.6 & 57.0 & 45.1 & 26.1 & 44.2    \\
SW~\cite{SW}          & R50       & 48.5 & 55.8 & 44.9 & 26.1 & 43.8    \\
DRPC~\cite{DRPC}        & R50       & 49.9 & 56.3 & 45.6 & 26.6 & 44.6    \\
GTR~\cite{GTR}         & R50       & 50.8 & 57.2 & 45.8 & 26.5 & 45.0    \\
ISW~\cite{ISW}         & R50       & 50.7 & 58.6 & 45   & 26.2 & 45.1    \\
SAN-SAW~\cite{SAN}     & R50       & 53.0 & 59.8 & 47.3 & 28.3 & 47.1    \\ 
Mask2former~\cite{mask2former} & R50       & 46.8 & 61.6 & 48.0 & 31.2 & 46.9    \\
HGFormer (\textbf{ours})  & R50       & 51.5 & 61.6 & 50.4 & 30.1 & \textbf{48.4}   \\ \hline
% IBN~\cite{IBN}         & R101      & 50.2 & 58.4 & 46.3 & 27.6 & 45.6    \\
% SW~\cite{SW}          & R101      & 50.1 & 56.2 & 45.2 & 27.2 & 44.7    \\
% DRPC~\cite{DRPC}        & R101      & 51.5 & 58.6 & 46.9 & 29.0 & 46.5    \\
% GTR~\cite{GTR}         & R101      & 51.7 & 58.4 & 46.8 & 29.1 & 46.5    \\
% ISW~\cite{ISW}         & R101      & 51.0 & 59.7 & 46.3 & 28.4 & 46.3    \\
% SAN-SAW~\cite{SAN}     & R101      & 54.7 & 61.3 & 48.8 & 30.2 & 48.8    \\
Mask2former~\cite{mask2former} & Swin-T & 51.3 & 65.3 & 50.6 & 34   & 50.3    \\
% Mask2former & Swin-T & 49.9 & 65.3 & 51.4 & 34.7 & 50.3    \\
HGFormer (\textbf{ours})    & Swin-T & 53.4 & 66.9 & 51.3 & 33.6 & \textbf{51.3}   \\ \hline
Mask2former\cite{mask2former} & Swin-L & 60.1 & 72.2 & 57.8 & 42.4        & 58.1    \\
HGFormer (\textbf{ours})  & Swin-L & 61.5 & 72.1 & 59.4 & 41.3        & \textbf{58.6}  \\
% \hline
\bottomrule
\end{tabular}
\label{tab:city2other}
}
\end{table}

\begin{table}[]
\centering
\caption{\textbf{Mapillary-to-other datasets generalization.} The models are trained on Mapillary, and tested on GTAV (G), Synthia (S), Cityscapes (C), and BDD (B).}
\vspace{-3mm}
% \small
\resizebox{\linewidth}{!}{
\begin{tabular}{c|c|ccccc}
% \hline
\toprule
         Method           & backbone  & G    & S    & C    & B    & Average       \\ \hline
IBN~\cite{IBN}                  & R50       & 30.7 & 27.0 & 42.8 & 31.0 & 32.9          \\
SW~\cite{SW}                   & R50       & 28.5 & 27.4 & 40.7 & 30.5 & 31.8          \\
DRPC~\cite{DRPC}                 & R50       & 33.0 & 29.6 & 46.2 & 32.9 & 35.4          \\
GTR~\cite{GTR}                  & R50       & 32.9 & 30.3 & 45.8 & 32.6 & 35.4          \\
ISW~\cite{ISW}                  & R50       & 33.4 & 30.2 & 46.4 & 32.6 & 35.6          \\
SAN-SAW~\cite{SAN}              & R50       & 34.0 & 31.6 & 48.7 & 34.6 & 37.2          \\ 
Mask2former~\cite{mask2former}          & R50       & 55.8 & 37.7 & 65.6 & 56.4 & 53.9          \\
HGFormer (\textbf{ours})    & R50       & 59.2 & 37.4 & 67.1 & 59.1 & \textbf{55.7}          \\  \hline
% IBN~\cite{IBN}                  & R101      & 30.7 & 27.0 & 42.8 & 31.0 & 32.9          \\
% SW~\cite{SW}                   & R101      & 28.5 & 27.4 & 40.7 & 30.5 & 31.8          \\
% DRPC~\cite{DRPC}                 & R101      & 33.0 & 29.6 & 46.2 & 32.9 & 35.4          \\
% GTR~\cite{GTR}                  & R101      & 32.9 & 30.3 & 45.8 & 32.6 & 35.4          \\
% ISW~\cite{ISW}                  & R101      & 33.4 & 30.2 & 46.4 & 32.6 & 35.6          \\
% SAN-SAW~\cite{SAN}              & R101      & 34.0 & 31.6 & 48.7 & 34.6 & 37.2          \\
Mask2former~\cite{mask2former}          & Swin-T & 57.8 & 40.1 & 68.2 & 59.1 & 56.3          \\
HGFormer (\textbf{ours})        & Swin-T & 60.1 & 39.5 & 69.3 & 61.0 & \textbf{57.5} \\ \hline
Mask2former~\cite{mask2former}          & Swin-L    & 64.8 & 48.4 & 77.9 & 64.7 & 63.9          \\
HGFormer (\textbf{ours})       & Swin-L    & 66.5 & 47.7 & 78.2 & 66.3 & \textbf{64.7}         \\
% \hline
\bottomrule
\end{tabular}\label{tab:mapillary}
}
\end{table}
\begin{table*}[]
\caption{\textbf{Cityscapes-to-Cityscapes-C generalization (level 5).}}
\vspace{-3mm}
\resizebox{\linewidth}{!}{
\begin{tabular}{c|c|cccc|cccc|cccc|cccc}
\toprule
\multirow{2}{*}{Method} & \multirow{2}{*}{Average} & \multicolumn{4}{c|}{Blur}      & \multicolumn{4}{c|}{Noise}   & \multicolumn{4}{c|}{Digital}  & \multicolumn{4}{c}{Weather} \\ \cline{3-18} 
                   &         & Motion & Defoc & Glass & Gauss & Gauss & Impul & Shot & Speck & Bright & Contr & Satur & JPEG & Snow & Spatt & Fog  & Frost \\ \hline
Mask2former-Swin-T~\cite{mask2former} & 41.6    & 51.5   & 49.4  & 38.2  & 46.2  & 9.6   & 9.8   & 13.5 & 44.4  & 74.2   & 60.0  & 70.0  & 23.3 & 23.7 & 59.4  & 65.4 & 27.3  \\
HGFormer-Swin-T (\textbf{ours})     & \textbf{43.9}  & 52.9   & 53.9  & 39.0  & 49.5  & 12.1  & 12.3  & 18.2 & 46.3  & 75.0   & 60.0  & 71.2  & 27.2 & 29.4 & 60.6  & 65.0 & 29.1  \\ \hline
Mask2former-Swin-L~\cite{mask2former} & 58.7    & 63.5   & 66.6  & 62.1  & 62.3  & 26.2  & 35.9  & 33.2 & 62.9  & 80.0   & 72.6  & 77.3  & 52.5 & 50.5 & 75.3  & 75.1 & 43.0  \\    
HGFormer-Swin-L (\textbf{ours})     & \textbf{59.4} & 64.1   & 67.2  & 61.5  & 63.6  & 27.2  & 35.7  & 32.9 & 63.1  & 79.9   & 72.9  & 78.0  & 53.6 & 55.4 & 75.8  & 75.5 & 43.2  \\
\bottomrule
\end{tabular}\label{tab:city_c_level5}
}
\end{table*}
\begin{table}[]
\centering
\caption{\textbf{Ablation of iterations in part-level mask classifcation.} In this ablation, HGFormer with Swin-T~\cite{swin} is trained on Cityscapes (C), and tested on Cityscapes, ACDC all (A), GTAV (G), BDD (B), Synthia (S), and Mapillary (M).}
\vspace{-3mm}
\resizebox{\linewidth}{!}{
\begin{tabular}{c|cccccc|c}
% \hline
\toprule
Iter & C    & A    & G    & B    & S    & M    & Avg  \\ \hline
1    & 76.8 & 56.1 & 51.3 & 52.1 & 32.1 & 65.8 & 55.7 \\
2    & 77.6 & 56.1 & 51.4 & 52.0 & 32.3 & 65.9 & 55.9 \\
3    & 77.9 & 56.2 & 51.8 & 52.6 & 32.8 & 66.2 & 56.2 \\
4    & 77.9 & 56.5 & 52.0 & 52.6 & 32.6 & 66.3 & \textbf{56.3} \\
5    & 77.8 & 56.4 & 51.7 & 52.6 & 32.5 & 66.3 & 56.2 \\
6    & 77.4 & 55.4 & 50.5 & 52.2 & 32.3 & 65.6 & 55.6 \\ 
% \hline
\bottomrule
\end{tabular}
\label{tab:iterations}
}
\end{table}

\begin{table*}[]
\centering
\caption{\textbf{Comparison of part-level classification and whole-level classification, and their combination.} We train HGFormer with Swin-T on Cityscapes, then test the model on other datasets.}
\vspace{-3mm}
\begin{tabular}{cc|cccccc}

\toprule
whole-level mask & part-level mask & ACDC (all) & GTAV  & BDD  & Synthia & Mapillary & Average \\ \hline
\checkmark      &                & 54.5       & 49.5 & 51.5 & 33.8    & 66.3      & 51.1    \\
                & \checkmark     & 56.2       & 51.3 & 53.1 & 33.3    & 66.5      & 52.1    \\
\checkmark      & \checkmark     & 56.6       & 51.3 & 53.4 & 33.6    & 66.9      & \textbf{52.4}   \\ 
\bottomrule
\end{tabular}
\label{tab:individualresult}
\end{table*}
% \begin{table}[]
% \centering
% \caption{\textbf{Cityscapes to Cityscapes-C generalization.}}
% \vspace{-3mm}
% \small
% \resizebox{\linewidth}{!}{
% \begin{tabular}{c|l|llll}
% \hline
% \multicolumn{1}{l|}{} & \multicolumn{1}{c|}{backbone} & \multicolumn{1}{c}{flops} & \multicolumn{1}{c}{City} & \multicolumn{1}{c}{City-C} & \multicolumn{1}{c}{Retention} \\ \hline
% FAN                   &                               &                           &                          &                            &                               \\
% Segformer             &                               &                           &                          &                            &                               \\
% Mask2former           &                               &                           &                          &                            &                               \\
% HGFormer              &                               &                           &                          &                            &                               \\ \hline
% \end{tabular}
% }
% \end{table}

\pparagraph{Mapillary-to-other datasets generalization.}
Here, models are trained on Mapillary~\cite{mapillary} and tested on BDD~\cite{BDD100k}, Mapillary~\cite{mapillary}, GTAV~\cite{GTAV}, and Synthia~\cite{synthia}. We can see that HGFormer is consistently better than Mask2former with all backbones, as shown in Tab.~\ref{tab:mapillary}. 
% HGFormer with R50 is 1.8 points better than Mask2former~\cite{mask2former}.

\pparagraph{Normal-to-corruption generalization.}
In this setting, models are trained on Cityscapes~\cite{cityscapes} and tested on Cityscapes-C~\cite{city_c} level 5, which includes 16 types of artificial corruptions at an extreme level. We compare HGFormer with Mask2former in Tab.~\ref{tab:city_c_level5}, showing that HGFormer is significantly better than Mask2former when generalizing to extremely corrupted images.

\begin{figure}[!t]
	\centering
	\vspace{-3mm}
	\includegraphics[width=0.94\linewidth]{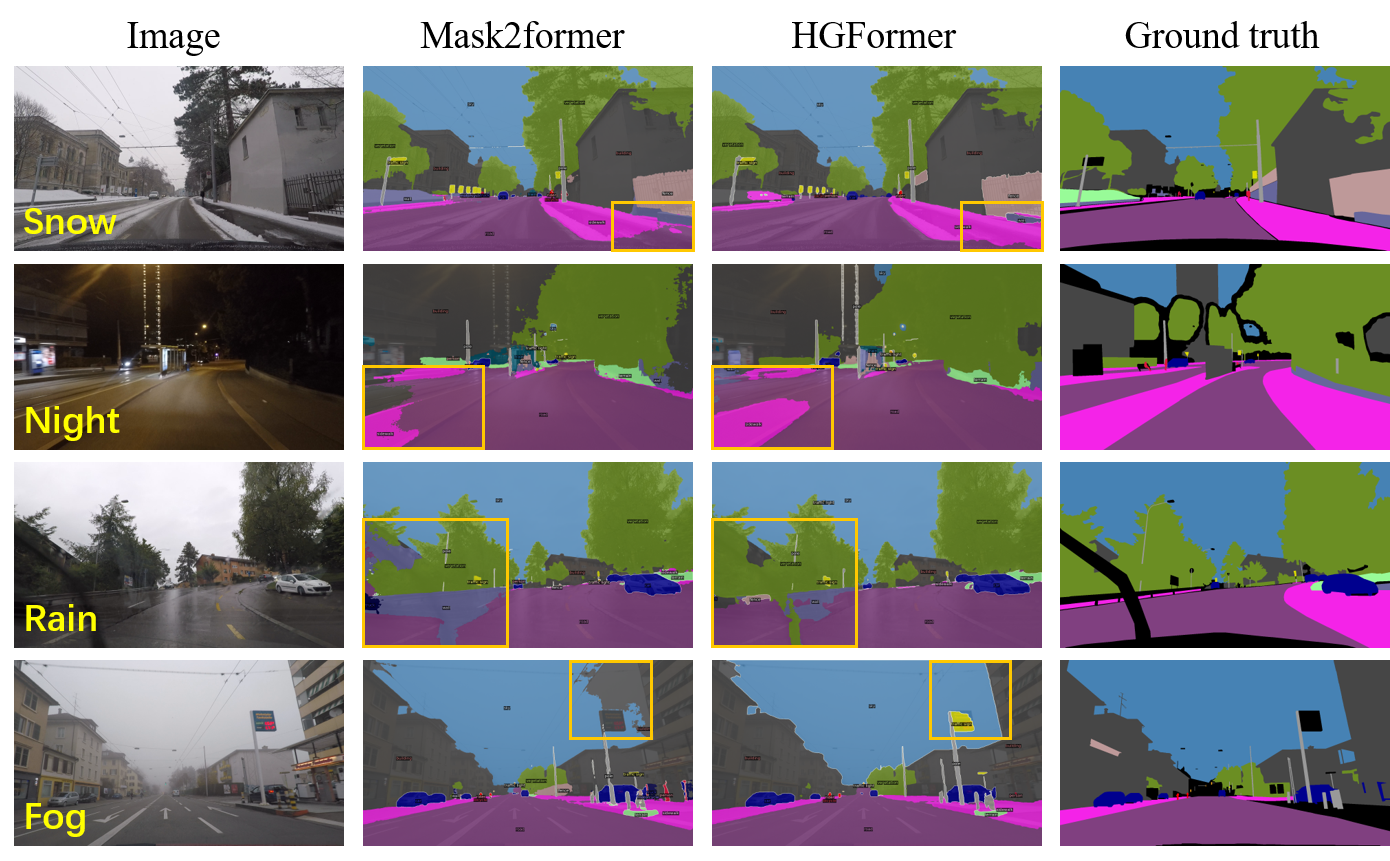}
	\vspace{-3mm}	
	\caption{\textbf{Visualization of results on adverse conditions.} The models are only trained on Cityscapes, and tested on images with adverse conditions. Our method is significantly better than the Mask2former~\cite{mask2former} under adverse conditions.
	}\label{fig:acdc}
\end{figure}
%  We adopt the recently proposed Mask2former~\cite{mask2former} as a strong baseline. 
\begin{figure*}[!t]
	\centering
% 	\vspace{-3mm}
	\includegraphics[width=0.85\linewidth]{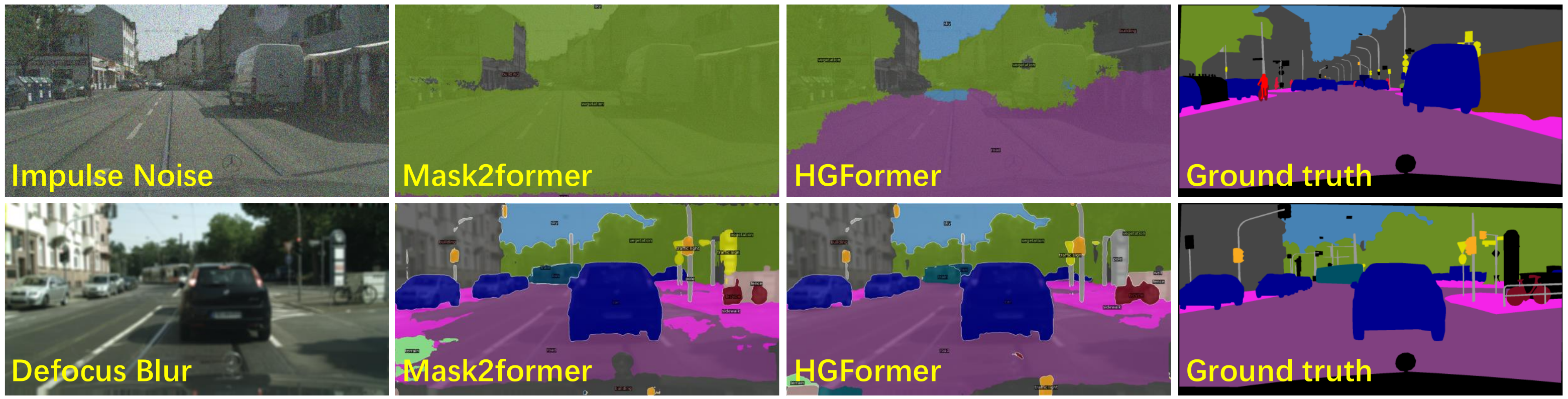}
	\vspace{-3mm}	
	\caption{\textbf{Visualization of results on corruptions.} We choose two kinds of corruption at level 5 for this visualization: impulse noise and defocus blur. The models are trained on Cityscapes. 
	}\label{fig:vislevel5}
\end{figure*}

\begin{figure*}[!t]
	\centering
	\vspace{-3mm}
	\includegraphics[width=0.85\linewidth]{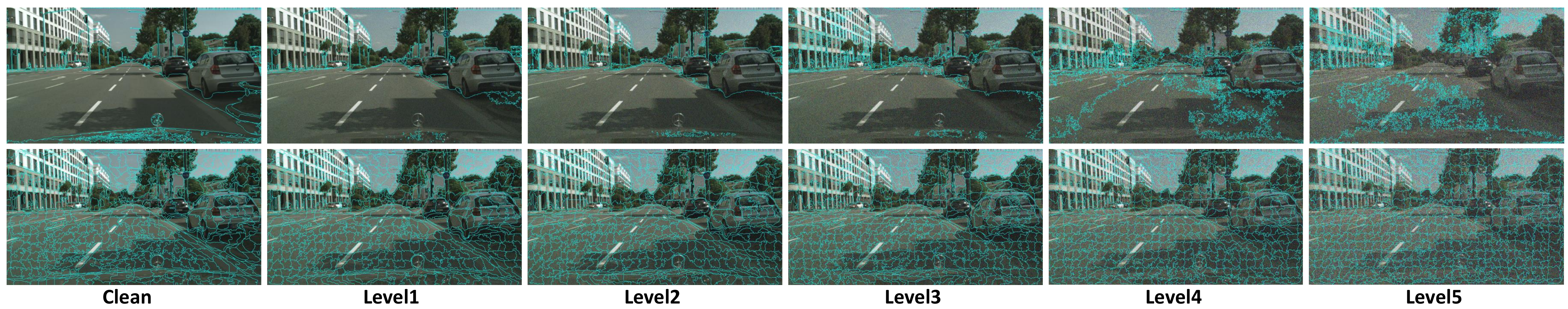}
	\vspace{-3mm}	
	\caption{\textbf{Visualization of part-level and whole-level masks at different levels of Gaussian noise.} In the first row, we visualize the whole-level masks from Mask2former. In the second row, we visualize the part-level masks from our method. We can see that our part-level masks are more robust than the whole-level masks in Mask2Former as the increasing severity level of Gaussian noise. }
	\label{fig:spixood}
\end{figure*}

\begin{figure*}[!t]
	\centering
	\vspace{-3mm}
	\includegraphics[width=0.84\linewidth]{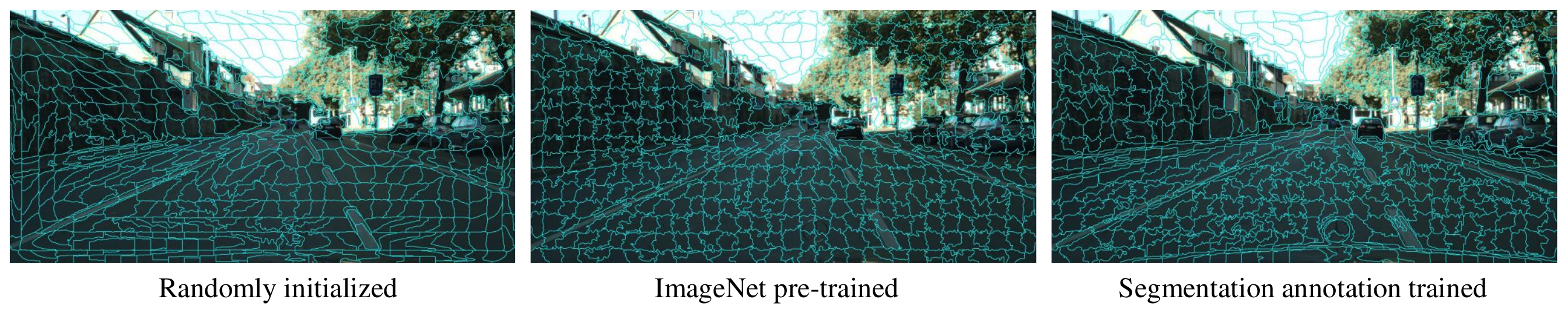}
	\vspace{-3mm}	
	\caption{\textbf{Visualization of part-level masks with different weights.} We find that the randomly initialized weights can also produce some reasonable part-level masks.}
	\label{fig:differentweights}
 	\vspace{-3mm}	
\end{figure*}

\subsection{Ablation Studies}
\pparagraph{Ablation of iterations in part-level mask classification.}
We test the results of different iterations of HGFormer and show the results in Tab.~\ref{tab:iterations}. It shows that the first iteration is much lower than later iterations on in-domain performance, and slightly lower than the later stages on out-of-distribution performance. As the iteration increases, the performance gradually increases. The performance is saturated at iteration 4. The last stage is lower than the second-last stage. We hypothesize that the last stage is influenced by the gradients from whole-level grouping since only the part-level tokens of the last stage are taken as the input of whole-level grouping. When we remove the whole-level grouping during training, the last stage is slightly higher than the second-last stage, which verifies our hypothesis.

\pparagraph{Individual performance of part-level and whole-level masks.}
We report the individual generalization performance of part-level and whole-level mask classification in Tab.~\ref{tab:individualresult}. It shows that the part-level classification is significantly better than the whole-level classification in HGFormer. And the ensemble of part-level and whole-level classification can further improve the performance, which indicates that the part-level and whole-level mask classification are complementary to each other.

% \pparagraph{Ensemble of plain mask classifications}
% Since each stage of \textit{plain grouping} can also produce individual semantic segmentation results, we report the ensemble of mask classification of Mask2former~\cite{mask2former} in Tab.~\ref{}. We can see that the ensemble of mask classification in Mask2former does not improve the robustness. We hypothesis that the diversity between the masks in the \textit{same level} is very low. 

% \pparagraph{Computational cost analysis}
% \begin{table}[]
% \begin{tabular}{ccc}
%             & \#Param (M) & GFLOPs \\
% Mask2Former & 44         & 529.9  \\
% HGFormer    & 48.18      & 512.9 
% \end{tabular}
% \end{table}
\subsection{Visualization Analyses}
% TODO: (1) diversity prediction of different formulations; 
% (2) comparison with mask2former and deeplabv3
% initial masks, just use imagenet pretrained weight, without any segmentation annotations
% (3) segmentation of new classes, has the potential for unseen class segmentation
% (4) part-level masks in unseen domain
\pparagraph{Visualization comparisons with Mask2former.}
We present the visualization results of Mask2forme and HGFormer, both with Swin-Tiny on ACDC (see Fig.~\ref{fig:acdc}) and Cityscapes-C (see Fig.~\ref{fig:vislevel5}) to demonstrate the performance of models for real-world adverse conditions and for synthetic corruptions. We choose impulse noise and defocus blur at level 5 for visualization. The results on both of the datasets show that HGFormer makes fewer errors than Mask2former in adverse conditions.

\pparagraph{Masks at different levels of corruption.}
 We visualize the part-level and whole-level masks at different levels of corruption to show how they change as the severity level increases (see Fig.~\ref{fig:spixood}). We can see that the whole-level masks are not stable with the increasing of severity levels, and totally failed at level 5. In contrast, the part-level masks are more stable, and can achieve high recall for the boundaries between classes.

\pparagraph{Part-level masks with different model weights.}
To provide more insights about our method, we visualize the part-level masks with model weights from random initialization, ImageNet pre-trained weights and Cityscapes trained weights. The results are shown in Fig.~\ref{fig:differentweights}. We can see that even using the randomly initialized weights and ImageNet pre-trained weights, our model can produce reasonable part-level masks, which indicates that our model has the potential for unsupervised segmentation and weakly-supervised segmentation. The results indicate that the \textit{part-level grouping structure itself can provide a good prior}, which can explain the generalization from the grouping side.  For the Cityscapes trained model weights, the \textit{boundaries between different categories} are more accurate than the randomly initialized and ImageNet pre-trained weights. It is worthwhile noting that the \textit{boundaries between the same class} is not unique, due to no ground truths being used for part-level masks. But with a part-level classification, all feasible part-level partitioning can be transformed to correct semantic segmentation results, if the \textit{boundaries between different classes} are correct.

\section{Conclusion and Future work}
In this paper, we propose a hierarchical semantic segmentation model, which can efficiently generate image partitions in a hierarchical structure. Then we perform both part-level mask classification and whole-level mask classification. The final semantic segmentation result is an ensemble of two results. Our method is verified to be robust in out-of-distribution images. We can explore more complicated fusion methods of classification at different scales. We leave them as future works. Since our model can be considered a kind of multi-task learning, how to automatically balance the loss weights of classification at different scales can be studied in the future.

\paragraph{Acknowledgement.} This work was supported by the National Nature Science Foundation of China under grants U22B2011 and 62101390. Jian Ding was also supported by the China Scholarship Council. We would like to thank Ahmed Abbas for the insightful discussions. 

%%%%%%%%% REFERENCES
{\small
\bibliographystyle{ieee_fullname}
\bibliography{egbib}
}

\end{document}